\documentclass[a4paper,conference]{IEEEtran}

\IEEEoverridecommandlockouts
\usepackage{cite}

\ifCLASSINFOpdf
\else
\fi
\usepackage{amsmath}
\usepackage{mathtools}
\usepackage{amsfonts}
\usepackage{xcolor}
\usepackage{textcomp}
\usepackage{multirow}

\usepackage{stfloats}
\begin{document}
%
\title{Inner Eye Canthus Localization for \\ Human Body Temperature Screening}

\author{
	\IEEEauthorblockN{
		Claudio Ferrari\IEEEauthorrefmark{1}, Lorenzo Berlincioni\IEEEauthorrefmark{1}\thanks{\IEEEauthorrefmark{1}Both authors contributed equally to this research.}, Marco Bertini, Alberto Del Bimbo
	}
	\IEEEauthorblockA{
		Media Integration and Communication Center\\
		University of Florence, Italy\\
		Email: \{claudio.ferrari, lorenzo.berlincioni, marco.bertini, alberto.delbimbo\}@unifi.it
	}
}

\maketitle

\begin{abstract}
In this paper, we propose an automatic approach for localizing the inner eye canthus in thermal face images. We first coarsely detect 5 facial keypoints corresponding to the center of the eyes, the nosetip and the ears. Then we compute a sparse 2D-3D points correspondence using a 3D Morphable Face Model (3DMM). This correspondence is used to project the entire 3D face onto the image, and subsequently locate the inner eye canthus. Detecting this location allows to obtain the most precise body temperature measurement for a person using a thermal camera. We evaluated the approach on a thermal face dataset provided with manually annotated landmarks. However, such manual annotations are normally conceived to identify facial parts such as eyes, nose and mouth, and are not specifically tailored for localizing the eye canthus region. As additional contribution, we enrich the original dataset by using the annotated landmarks to deform and project the 3DMM onto the images. Then, by manually selecting a small region corresponding to the eye canthus, we enrich the dataset with additional annotations. By using the manual landmarks, we ensure the correctness of the 3DMM projection, which can be used as ground-truth for future evaluations. Moreover, we supply the dataset with the 3D head poses and per-point visibility masks for detecting self-occlusions. The data will be publicly released.  
\end{abstract}


%
\IEEEpeerreviewmaketitle

\section{Introduction}\label{sec:intro}
When properly implemented, infrared thermography is the most efficient non-contact, cost effective and reasonably accurate solution for mass screening of individuals for elevated body temperature; this type of screenings is commonly used in public environments like airports, malls, hospitals and has become a way to non intrusively detect body fever, a common precursor of diseases like H1N1 flu, SARS, MERS and COVID-19 \cite{WILDERSMITH2020e102, bitar2009international, NG2004104}. Such screening is normally conducted by measuring the skin temperature on a subject's face. However, not all the face regions are suitable for this task; it is widely recognized that the inner eye canthus represents the most stable region where to obtain a reasonably reliable temperature measure \cite{ring-2010, mercer2009fever}. 
This because, in normal conditions, it is both the warmest face region, and the most invariant to environmental factors that can alter the skin superficial temperature and, unlike the forehead, is not influenced by exertion and physical activity. 

However, accurately detecting such regions on thermal imagery is a very challenging task as: (i) even slight head pose changes will result in self-occlusions and consequent obstruction of the canthus; (ii) the majority of thermal face databases do not contain annotations or, in case they do, they do not explicitly account for those regions, making it hard to both develop and evaluate a detection algorithm. 

In this paper we propose a method for automatic detection of the inner eye canthus region based on coarse detection of 5 facial keypoints corresponding to eyes, nose and ears in thermal images, and on their 2D-3D mapping on a 3D morphable face model (3DMM). This process  has several benefits over direct region localization, since it allows to obtain additional information regarding the pose, i.e.~understanding if the face is directly looking at the camera and estimating self-occlusions of face parts.

A second contribution of this work is the release of additional face region annotations for a thermal face images dataset, including 3D head pose and face regions visibility masks.

The paper is organized as follows: related works are discussed in Sect.~\ref{sec:related}, the proposed method is described in Sect.~\ref{sec:proposed}; the dataset used and the additional annotations created to extend it are presented in Sect.~\ref{sec:dataset}, while experimental results are presented in Sect.~\ref{sec:exp-res}. Finally conclusions are drawn in Sect.~\ref{sec:conclusions}.

\section{Related Work}\label{sec:related}
The scientific literature on visible spectrum face images is extremely vast; even considering only works related to facial landmarks detection and pose recognition it is better to refer the reader to surveys like \cite{JIN20171, Wu:2019aa}.
In recent years, the improvement in resolutions and decrease in costs of thermal cameras has led to the development of methods specifically designed for the analysis of images in this spectrum.

\paragraph*{Thermal face datasets}
Due to the difficulty in creating them, compared to visible spectrum datasets, thermal face datasets are much smaller both in terms of number of identities (typically a few tens) and images (typically a few hundreds or thousands).

The PUCV Thermal Temporal Face (PUCV-TTF) dataset \cite{Hermosilla-2017} consists of thermal images taken over time, so to study temporal changes in thermal imaging. It includes 46 people with five subsets for each subject, and each subset has 50 images.

The UND Collection X1 dataset \cite{nd-x1} is composed of $\sim 2300$ images of 82 subjects, and contains both high resolution visible images and corresponding low resolution ($320 \times 240$ pixels) thermal images.

The UL-FMTV dataset \cite{Ghiass-2018} of high resolution thermal videos has been obtained from 238 subjects over four years, considering pose, occlusions, time lapse.

In \cite{kopaczka2018fully} a high-resolution thermal dataset of faces, filmed in different poses and expressions has been provided; it is composed of $\sim 3000$ images from 90 subjects. Faces have been manually annotated with 68 keypoints.

\paragraph*{Thermal image analysis}
In general, thermal images have proven a challenging obstacle for many computer vision and image analysis tasks, like face recognition and re-identification \cite{ghiass2014infrared}, face part detection, keypoint localization, etc., due to their low contrast, low resolution, and lack of texture information.

In \cite{wu2016thermal} has been presented one of the first CNN architectures to perform face recognition in the thermal domain only; in \cite{Hermosilla-2017} local engineered features like SIFT, SURF, LBP have been used to evaluate the performance of face recognition under varying temporal conditions that affect thermal imaging, like environmental conditions and physiological changes that may happen over time. 
Face recognition can be hampered by the presence of glasses, that result in facial occlusion in thermal imagery; to deal with such occlusions the method presented in \cite{Kumar-2020} uses a bag of CNN features model.

In \cite{Poster_2019_CVPR_Workshops} the authors evaluate how different deep neural network architectures for keypoint localization adapt to thermal face imaging, in order to perform thermal-to-visible face alignment and recognition. The results show that learning a global face appearance reduces critical localization errors.

In \cite{Kopaczka-2019} the authors of the dataset originally proposed in \cite{kopaczka2018fully} evaluate the performance of different keypoint localization with deep neural network comparing this approach with SIFT and HOG, then perform face expression recognition. The same authors have proposed in \cite{Kopaczka-2018b} a modular system for face detection, face tracking, head pose estimation, and emotion recognition using HOG and SIFT features.

In \cite{chu-2019} is proposed to use a multi-task neural network to jointly consider facial landmark detection and emotion recognition for thermal face images. The network is composed by two parts: the first one is based on the U-Net structure, to extract good features, the second part of the network contains two branches, one for landmark detection and the other for emotion recognition. 

\begin{figure*}[!t]
\centering
\includegraphics[width=0.99\linewidth]{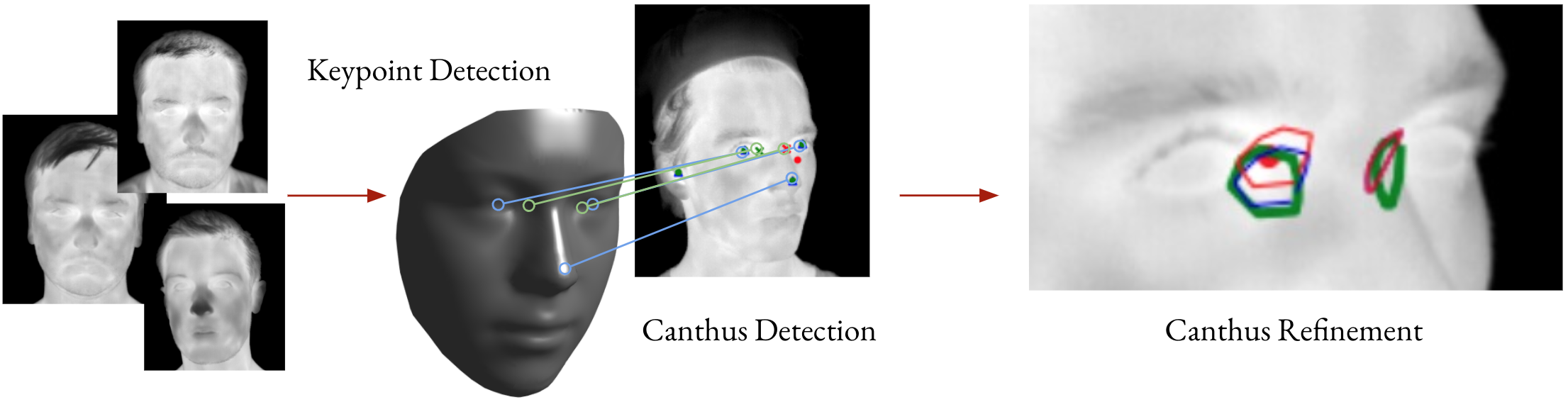}
\caption{Proposed method overview. Given a thermal face image, we first detect 5 keypoints (center of the eyes, ears, and nosetip) using the OpenPose detector~\cite{openpose}; then, we employ a 3DMM~\cite{Ferrari:2017b} to estimate the head pose and project the 3D face model on the image plane exploiting 2D-3D correspondence of points (blue lines). We subsequently locate the eye canthi by means of the 3D model (green lines). Finally, ( we refine the canthi localization (red point) by searching for the warmest area around the estimated canthi (green circles).}
\label{fig:pipeline}
\end{figure*}

\paragraph*{Thermography}
Face thermography is a useful new technique for psychophysiological research and medical applications; unlike other physiological measures it is uniquely contact-free, allowing its deployment in many real-world scenarios, like airports, hospitals, etc.

In \cite{vardasca2017influence} the authors have analyzed the influence of angles and distance on assessing temperature measurement using inner-canthi of the eye and thermal cameras, finding that when the face is not directly facing the camera, e.g.~rotating more than 30\textdegree\ may result in a measure difference of 1-2\textdegree C.

In \cite{kopaczka2018towards} the authors propose to use variation in temperature measurements of different regions of the face to recognize the presence of mental stress; the authors find that signals extracted from the upper lip region correspond well with high stress levels, while no correspondence can be shown for the other regions like forehead and nose. Use of thermographic measurement from faces of students has been proposed to assess learning difficulty in digital learning environments in \cite{Namrata-2019}. 
In \cite{bitar2009international} the authors use both a task-constrained deep convolutional network pretrained to detect 5 keypoints and then to detect 68 keypoints, in combination with OpenPose, to recognize face landmarks in thermal images. These landmarks are then used to identify 4 ROIs (forehead, nose tip, right and left cheeks) used to measure skin temperature, in order to detect temperature changes associated with auditory stimulus.

Fever screening using termography has received a large attention following previous pandemic outbreaks like the severe acute respiratory syndrome (SARS) \cite{Chan-2006} and H1N1 flu (also known as ``swine-flu") \cite{mercer2009fever}. In these works the best correspondences between face parts and core body temperature have been analyzed resulting in guidelines and international standards on how to perform the screening, e.g.~selecting the inner canthi area of the eyes \cite{mercer2009fever, Pascoe-2010}.

In order to automatize the process a few works have dealt with eye localization in thermal images.
In \cite{Budzan-2013} the Randomized Hough Transform has been used to localize the ellipses of the eyes and thus their inner canthi. The method has been tested on a dataset of 125 faces with a maximum angle deviation of 15\textdegree\ from the frontal pose. In \cite{Lin_2019_ICCV} continuous body temperature measurements is obtained using a simple geometric model to detect forehead in thermal images and fine tuning a SSD detector with MobileNet backbone to perform face detection in thermal images.
In \cite{Aryal-2019} RGB and thermal imagery is combined to compute face landmarks in the visible spectrum and then obtain thermometry information with the thermal camera, to assess comfortable indoor conditions.
Thermometric information obtained from a set of keypoints localized next to veins and capillaries has been proposed in \cite{Lin-2019} to perform face recognition.

\section{Proposed Method}\label{sec:proposed}
The proposed algorithm consists of three main steps:
\begin{itemize}
    \item Coarse detection of 5 keypoints \textit{i.e.} eyes, nosetip and ears;
    \item 3D head pose estimation, model projection, and localization of the inner eye canthus region;
    \item Eye canthus location refinement.
\end{itemize}
\noindent
The 5 initial keypoints are detected by means of an ``off the shelf'' detector \textit{i.e.} OpenPose (OP)~\cite{openpose}. Despite being designed for visible spectrum imagery, it can robustly detect body, face, and hands keypoints rather accurately also on thermal images. In this work, we employ only the 5 facial keypoints, discarding all the others. As 3D face model, we employ the DL-3DMM proposed by Ferrari \textit{et al.}~\cite{Ferrari:2017b}. The DL-3DMM comprises a generic 3D face model $\mathbf{m} \in \mathbb{R}^{3 \times 6704}$ with 6704 vertices, and a set of deformation components $\mathbf{C} \in \mathbb{R}^{300 \times 6704}$ that allow deforming the generic model.

In the following, we separately describe the two subsequent steps \textit{i.e.} pose estimation plus model projection, and the eye canthus localization refinement. In Fig.~\ref{fig:pipeline}, the whole framework is illustrated.

\subsection{Head Pose Estimation and Model Projection}\label{subsec:3d-head-pose}
In order to estimate the 3D pose and a 3D to 2D projection exploiting the OP keypoints, we need to annotate a corresponding set of points on the 3D model. Let $\mathbf{l}_{op} \in \mathbb{R}^{2 \times 5}$ and $\mathbf{L}_{op} \in \mathbb{R}^{3 \times 5}$ be the 2D detected and 3D annotated keypoints, respectively. We estimate the projection using the orthographic camera model:

\begin{equation}
\label{eq:simTrans}
\mathbf{l}_{op} =  \mathbf{A}\cdot\mathbf{L}_{op} + \mathbf{t} \; ,
\end{equation}
\noindent
 where $\mathbf{A} \in \mathbb{R}^{2 \times 3}$ contains the affine camera parameters, and $\mathbf{t} \in \mathbb{R}^{2\times5}$ is the translation on the image. To estimate the parameters, firstly, we recover the affine matrix $\mathbf{A}$ by solving the following least squares problem: 
 
 \begin{equation}
\underset{\mathbf{A}}{\arg\min} \left \| \mathbf{l}_{op} - \mathbf{A} \cdot \mathbf{L}_{op} \right \|_{2}^{2}\; ,
\end{equation}
\noindent
for which the solution is given by $\mathbf{A} = \mathbf{l}_{op} \cdot \mathbf{L}_{op}^{+}$, where $\mathbf{L}_{op}^{+}$ is the pseudo-inverse matrix of $\mathbf{L}_{op}$. We can employ a simple least squares solution since, by construction, OpenPose assumes a consistent structure of the 3D face parts, so not permitting unreasonable keypoint arrangements (e.g., the nose will never be detected above the eyes).
Finally, the 2D translation can be estimated as $\mathbf{t} = \mathbf{l}_{op}-\mathbf{A}\cdot\mathbf{L}_{op}$. The estimated projection $\mathbf{P} = [\mathbf{A}, \mathbf{t}]$ is used to map each vertex of the model $\mathbf{m}$ onto the image.

As discussed previously, even slight head rotations can lead to the self-occlusion of the eye canthus regions, being them located in the nose side concavity. Thus, in order to be able to accurately perform a reliable temperature measurement at that points, we need to detect whether they are visible o not, totally or even partially. To this aim, we extract the 3D rotation matrix $\mathbf{R} \in \mathbb{R}^{3 \times 3}$ from the affine projection matrix  $\mathbf{A}$ by means of QR decomposition. This is possible thanks to the orthogonality property of rotation matrices \textit{i.e.} $\mathbf{R}^T = \mathbf{R}^{-1}$. We then use the rotation $\mathbf{R}$ to rotate $\mathbf{m}$ according to the estimated pose, and calculate the visibility of each 3D vertex from the novel vewpoint using the method proposed by Katz \textit{et al.}~\cite{katz2007direct}. This way, we can easily detect which face parts are self-occluded.  

\begin{figure}[!b]
\centering
\includegraphics[width=0.94\linewidth]{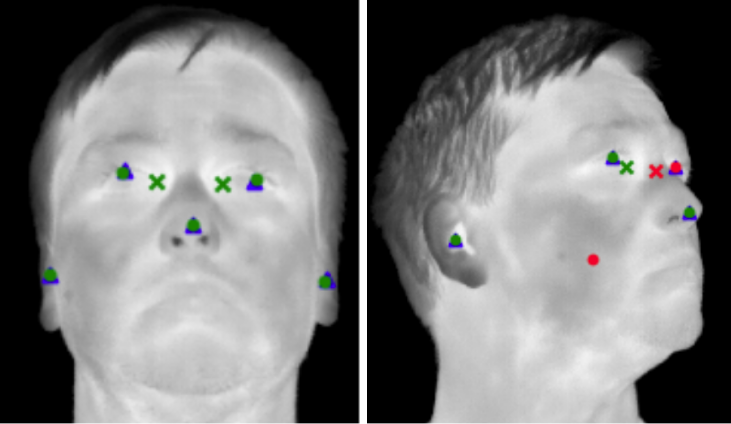}
\caption{Visualization of OpenPose keypoints (blue triangles) and our estimated projections (green/red dots). Red color indicates the projected keypoint is marked as occluded. The green crosses are our projected eye canthus points, which are not detected by OpenPose.}
\label{fig:projection}
\end{figure}

The eye canthus region, as well as any other face point, can now be detected on the face image first by annotating the points of interest in the 3D model, and subsequently calculating their corresponding 2D coordinates by means of the projection matrix $\mathbf{A}$. Note that the annotations must be done just once. An example is shown in Fig.~\ref{fig:projection}; the green crosses represent the projection of the eye canthus that were manually annotated on the 3D model and projected onto the image. Note also that when the subject is not frontal, we can detect the self occlusion. This aspect is important since OpenPose (as well as the majority of landmark detectors) do not usually provide indication of occlusions; rather, they either attempt to approximate the localization even if points are not directly visible, or do not return the detection at all. 

\subsection{Eye Canthus Refinement}\label{subsec:canthus-ref}
\begin{figure}[!t]
\centering
\includegraphics[width=0.99\linewidth]{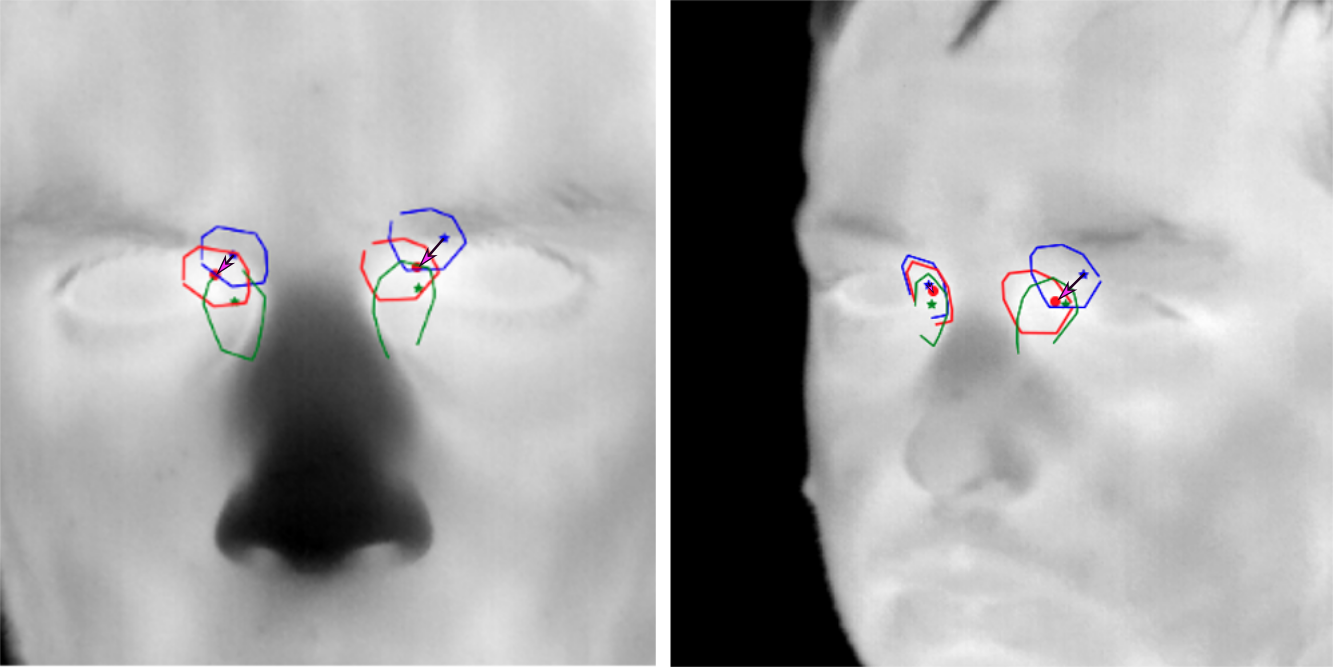}
\caption{Examples of eye canthus region refinement. In green manually annotated chantus and region, in blue OpenPose estimation and in red the refinement result.}
\label{fig:refinement}
\end{figure}

One limitation of the proposed approach is that it heavily relies on the accuracy of the keypoint detections. In order to account for possible inaccuracies, we refine the canthus localization by searching for the warmest point in a neighborhood. This is motivated by the assumption that, locally, the eye canthus and its surrounding is expected to be the warmest face area~\cite{mercer2009fever}. So, we first annotate the two eye canthus $\mathbf{L}_{ch}$ in the 3D model $\mathbf{m}$, and define a region around them;  each region is defined as the convex hull of the k-ring of $\mathbf{L}_{ch}$ (k-ring is the set of vertices that are distant from $\mathbf{L}_{ch}$ at most $k$). Within this area, we take the hottest point, \textit{i.e.}~the brightest pixel, and consider it as the new center of the eye canthus. An example of this process is shown in Fig.~\ref{fig:refinement}. Prior to computing the hottest point, we perform a Gaussian smoothing to account for sensor noise.

This strategy has a two-fold advantage: first, it allows correcting possible inaccurate detections; second, it ensures the measurement of the maximum temperature, representing at all effect a safe upper bound. In terms of thermal screening, this allows preventing dangerous false negatives.

\section{Validation Data}\label{sec:dataset}
\begin{figure}[!t]
\centering
\includegraphics[width=0.95\linewidth]{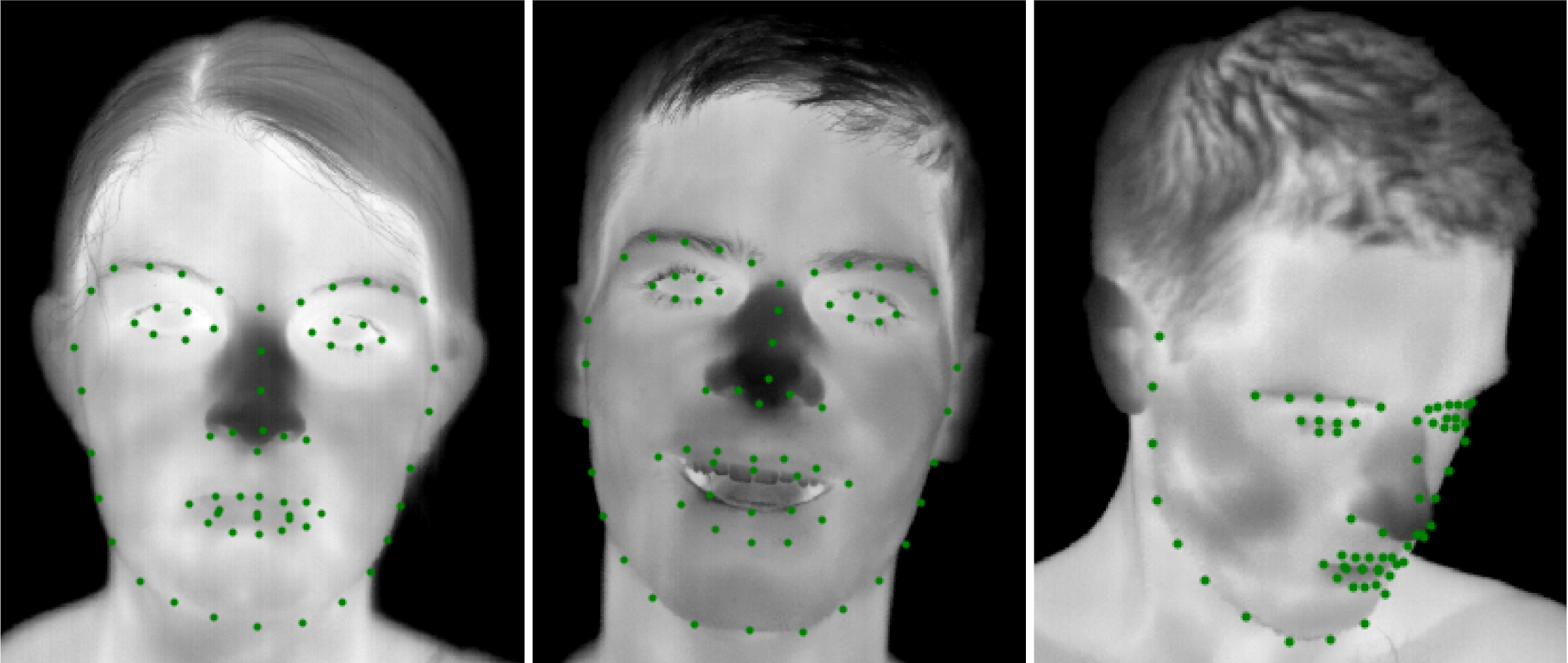}
\caption{Examples of the manually annotated landmarks in the dataset~\cite{kopaczka2018fully}.}
\label{fig:manualLm}
\end{figure}

In this section, we describe the dataset used for validating our approach, and how we enriched such data to allow quantitatively assessing the localization accuracy. We make use of the FaceDB dataset collected by Kopaczka \textit{et al.}~\cite{kopaczka2018fully}. It contains 2935 high resolution ($1024 \times 768$ pixels) thermal frames of 90 subjects; for each subject, various sequences are recorded including pose variations, expression variations and basic action units activation. This dataset is one of the few including manually annotated landmarks; more precisely, each face image is annotated with 68 landmarks following the standard configuration as used in other datasets, such as the Helen dataset~\cite{le2012interactive} (see Fig.~\ref{fig:manualLm}). In this particular configuration, some of those landmarks do cover the contour of the eyes, including the eye canthus; however, we argue that, if not properly instructed to locate the canthus, different persons could interpret its location slightly differently. Moreover, as previously discussed, the ISO/TR 13154 termograph screening standard guidelines suggest to perform the measurement considering a small region rather than a single point, as reported in~\cite{mercer2009fever}. Thus, it is important to correctly identify such region, and ground-truth annotations are missing in this case. 

To address this, we propose to use our solution to augment the dataset and provide additional ground-truth annotations, that are derived automatically. In particular, we make use of the manually annotated landmarks to:

\begin{itemize}
    \item Estimate the 3D pose and project the 3D model;
    \item Deform the 3D model to more accurately fit the face images, accounting for facial expressions;
    \item Identify the eye canthus regions by manually annotating a set of points in the 3D model and localizing the corresponding area on the images by means of the estimated projection.
\end{itemize}

The first step is performed as described in Section~\ref{subsec:3d-head-pose}, this time considering all the 68 available landmarks, which are assumed to be all correct and accurate. Next, we deform the 3DMM to fit the landmark locations, thus adapting its shape to each face image. While this is not possible with the OpenPose detections, that includes only 5 keypoints, the available 68 landmarks are sufficient to obtain a reasonable approximation of the face surface (see Fig.~\ref{fig:manualLm}). To this aim, we exploit the method proposed in~\cite{Ferrari3DV}, which is fast and can account for strong facial expressions. Such method performs a geometric-based 3DMM fitting; it deforms the generic model $\mathbf{m}$ with the goal of minimizing the reprojection error between the annotated 2D landmarks and those obtained by projecting the 3D landmarks onto the image. To this aim, we estimate the 3DMM deformation coefficients $\boldsymbol{\alpha}$ solving the following:

\begin{equation}
\label{eq:sparse-coding-b}
\small
\underset{\boldsymbol{\alpha}}{\min}
\left \|\mathbf{l}_{gt} - \mathbf{P}(\mathbf{L}_{gt} - \mathbf{C}\boldsymbol{\alpha} ) \right \|_{2}^{2} + \lambda \left \| \boldsymbol{\alpha} \circ \boldsymbol{\mathbf{\mu}}^{-1} \right \|_{2} \; .
\end{equation}
\noindent
where $\mathbf{l}_{gt}$ and $\mathbf{L}_{gt}$ are the 2D ground-truth landmarks and the 3D landmarks respectively, $\mathbf{P}$ indicates the projection onto the image, and $\boldsymbol{\mathbf{\mu}}$ is a regularization term. To solve the problem we first pre-compute the landmarks displacement $\mathbf{X} = \mathbf{l}_{gt} - \mathbf{P}\mathbf{L}_{gt}$ and the deformation components projection $\mathbf{Y} = \mathbf{P}\mathbf{C}$. The latter is necessary to be able to fit the landmark locations directly on the image plane, The solution can be then found in closed form as:

\begin{equation}
\label{eq:def-coefficients-estimation}
\boldsymbol{\alpha} = \left (\mathbf{Y}^T \mathbf{Y} + \lambda \cdot \text{diag}(\hat{\boldsymbol{\mu}}^{-1}) \right)^{-1}\mathbf{Y}^T\mathbf{X} \; ,
\end{equation}
\noindent
For more details on the solution of the minimization problem, the reader can refer to~\cite{Ferrari:2017b}. The average model $\mathbf{m}$ is then deformed into a new shape $\mathbf{S}$ as follows:

\begin{equation}
\label{eq:3dmmL}
\mathbf{S} = \mathbf{m} + \sum_{i=1}^{k} \mathbf{C}_i \alpha_i \; .
\end{equation}
\noindent
This step is fundamental to obtain a consistent annotation across the images. In fact, if not properly modeled, facial expressions would eventually induce a misalignment between the reprojected points and compromise the pose estimation accuracy. The deformed 3D model $\mathbf{S}$ is then projected onto the image and used to annotate the eye canthi regions; in particular, similarly to Section~\ref{subsec:canthus-ref}, the convex hull defined by the k-ring of the eye canthus ($k = 1, \dots, 4$) is used as ground-truth.

Finally, using the same strategy as expounded in Section~\ref{subsec:3d-head-pose}, we estimate a dense visibility map for each deformed model $\mathbf{S}$, so that each image of the dataset is enriched with such information.

\begin{figure}[!t]
\centering
\includegraphics[width=0.99\linewidth]{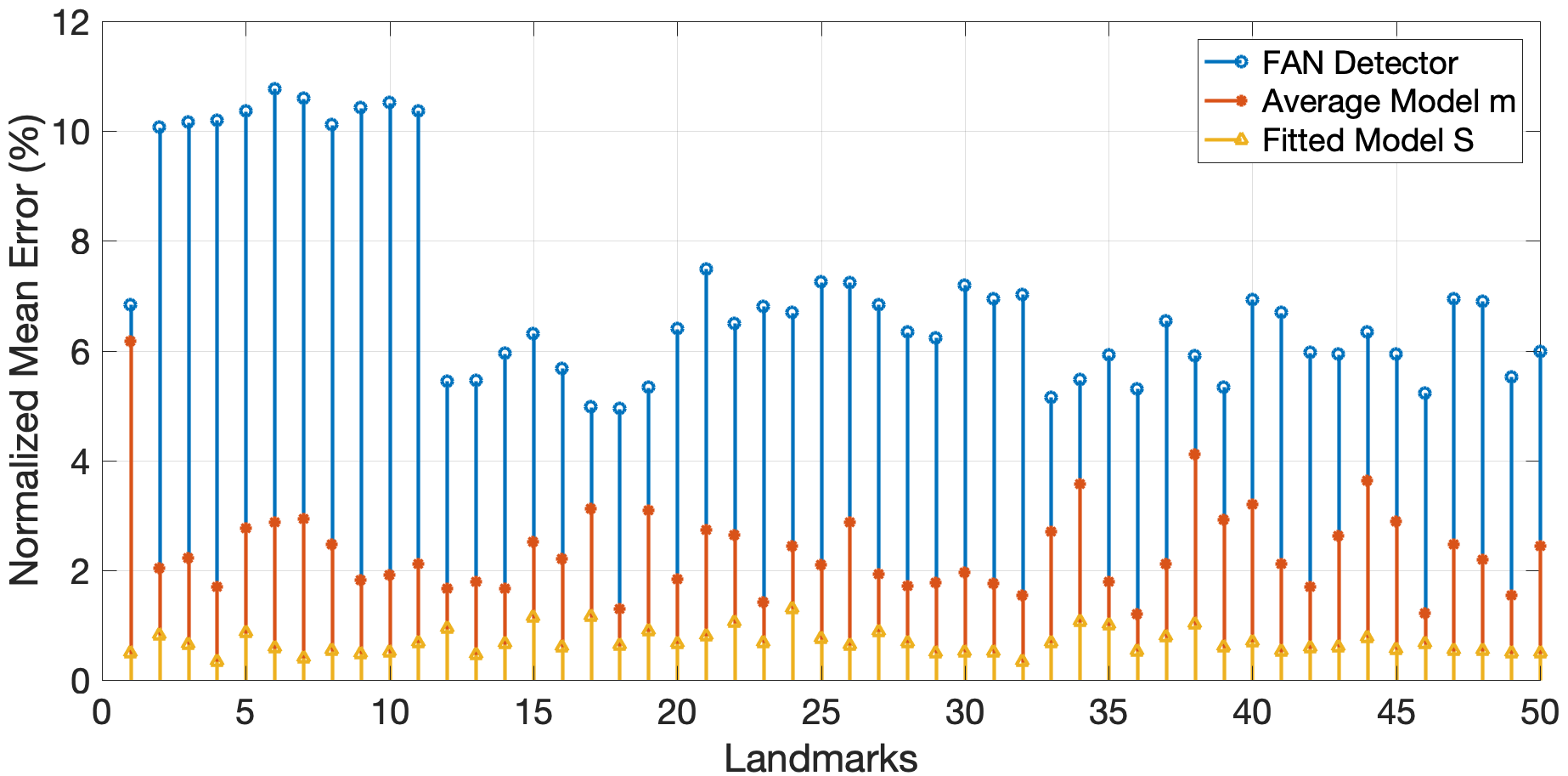}
\caption{Landmarks detection/reprojection error. In this configuration, the left/right eye canthi are the landmarks 22 and 25, respectively. }
\label{fig:lmError}
\end{figure}

\section{Experimental Results}\label{sec:exp-res}
We validate the proposed eye canthus detection approach with an extensive set of experiments. In this regard, we previously discussed that, for this particular task, performing a quantitative accuracy assessment is made difficult by the partial lack of ground-truth data. So, we first validate the new ground-truth data that we generated as expounded in Section~\ref{sec:dataset}. Then, using this new data, we report detection accuracy results for the proposed eye canthus detection. 

\begin{figure*}[!t]
\centering
\includegraphics[width=0.99\linewidth]{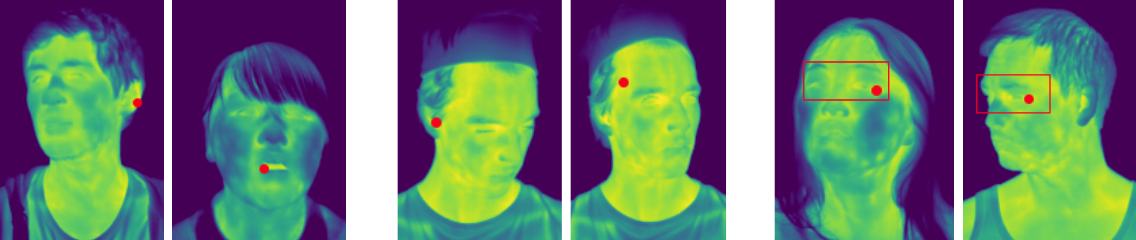}
\hspace{5.5cm}
(a)
\hspace{5.5cm}
(b)
\hspace{5.5cm}
(c)
\caption{Examples of wrong detections when searching for the hottest point. Wrong detections can be caused for example by environmental factors (a), or changes in pose that influence the skin measurement (b). Even if restricting the search to the eyes region (c) wrong detection still occur. }
\label{fig:misdetection}
\end{figure*}

\subsection{Ground-truth validation}\label{subsec:gt-validation-exp}
In this section, we provide a quantitative assessment of the additional ground-truth data. It is important that the new data which is intended to be used as ground-truth is correct. To this aim, we evaluate the landmarks reprojection error with respect to the manual annotations contained in the dataset, in order to ensure our annotations are meaningful. We also show that the 3DMM deformation step is important to account for the presence of facial expressions that can impair a correct pose and projection estimation. In Fig.~\ref{fig:lmError}, the Normalized Mean Error (NME) is reported separately for each landmark. The NME is a standard measure in the field of landmark detection and is computed as the Euclidean distance between the ground-truth and detected landmarks, normalized by the square root of the face bounding box size. Since no annotations are provided for such bounding box, we used an approximated box defined by the external face contour landmarks. Note that we do not use the latter landmarks for neither estimating the pose nor fitting the 3DMM; this because defining a unique face contour in 2D is not possible due to self-occlusions, which make the definition of the face silhouette inconsistent across 2D and 3D~\cite{qu122015adaptive}.

From Fig.~\ref{fig:lmError}, we can see that adapting the 3DMM is fundamental to obtain accurate model projections. We also reported the error obtained using the state-of-the-art FAN landmark detector~\cite{bulat2017far} as baseline. Compared to that, we argue that our annotations can be considered sufficiently accurate to be used as ground-truth for further evaluation.

\subsection{Eye canthus region detection}\label{subsec:canthus-exp}
We report here a quantitative evaluation of the eye canthus detection accuracy, performing three different experiments. We first evaluate the detection accuracy with respect to the eye canthus landmarks in terms of NME; second, we use our annotations to compute the intersection over union (IoU) between the ground-truth and detected canthus regions. Finally, we also use the additional visibility masks to assess the accuracy of the occlusion detection. Results are reported in Table~\ref{tab:eye-detect}. NME errors are computed both with respect to the manual annotations contained in the dataset (man) and our proposed annotations (gt); we can observe that the two measures are accurate and very similar, again confirming the validity of the proposed data. The IoU measure instead suggests that using a larger region can be convenient to better approximate the correct area. We can observe we also correctly detect whether the canthi are occluded or not in the 90\% of the cases. The occlusion detection is formulated as binary decision (occluded / non-occluded ) applied to both the eyes, using the estimated visibility map.

In all the cases, we witness a slight drop of accuracy after the warmest point-based refinement. We already discussed that manual annotators could have not been properly instructed to label the eye canthi for the particular goal of performing thermal screening, and this could be a result of that. However, in this particular case, looking for the warmest point is beneficial in as much as it prevents under-estimating the temperature, which could potentially lead to dangerous false negatives. 

\begin{table}[!t]
\renewcommand{\arraystretch}{1.1}
\caption{Eye Canthus detection accuracy (* means result is the same)}
\label{tab:eye-detect}
\centering
\scalebox{0.95}{
\begin{tabular}{|c|c|c|c|c|}
\hline
& & FAN~\cite{bulat2017far} & OP+3DMM & Refinement\\
\hline
\multirow{3}{*}{1-Ring} &
IoU  & -  & 16.5 $\pm$ 3.5 & 8.1 $\pm$ 2.1\\
& NME (man)(\%) & 6.8 * $\pm$ 2.1  & 4.1 $\pm$ 0.3 * & 5.6 $\pm$ 1.3 \\
& NME (gt)(\%) & 6.5 * $\pm$ 2.2 & 3.7 $\pm$ 0.3 * &  4.9 $\pm$ 1.3 \\
\hline
\multirow{3}{*}{2-Ring} &
IoU  & -  & 32.5 $\pm$ 4.8 & 23.4 $\pm$ 3.6 \\
& NME (man)(\%) & 6.8 *$\pm$  2.1  & 4.1 $\pm$ 0.3 * & 5.1 $\pm$ 1.1  \\
& NME (gt)(\%) & 6.5  *$\pm$ 2.2 & 3.7 $\pm$ 0.3 * &  4.5 $\pm$ 1.1 \\
\hline
\multirow{3}{*}{3-Ring} &
IoU & -  & 41.7 $\pm$ 4.5 &  34.6 $\pm$ 2.8 \\
& NME (man)(\%) & 6.8 *$\pm$  2.1  & 4.1 $\pm$ 0.3 * & 4.8 $\pm$  1.2 \\
& NME (gt)(\%) & 6.5  *$\pm$ 2.2 & 3.7 $\pm$ 0.3 * & 4.3 $\pm$ 1.1\\
\hline
\multirow{3}{*}{4-Ring} &
IoU  & -  & 47.1 $\pm$ 4.5 & 39.8 $\pm$ 2.8 \\
& NME (man)(\%) & 6.8 *$\pm$  2.1  & 4.1 $\pm$ 0.3 * & 4.8 $\pm$ 1.1 \\
& NME (gt)(\%) & 6.5  *$\pm$ 2.2 & 3.7 $\pm$ 0.3 * &  4.4 $\pm$ 1.1\\
\hline
& Occlusion (\%) & -  & 89.9  & - \\
\hline
\end{tabular}
}
\end{table}

\subsection{Head Pose Estimation}\label{subsec:headpose-exp}
We evaluate the accuracy of the head pose estimation in terms of head rotation angles (Euler angles), that is pitch angle $\alpha$ ($x$-axis), yaw angle $\beta$ ($y$-axis), and roll angle $\gamma$ ($z$-axis or in-plane rotation). The three angles are extracted from the rotation matrices $\mathbf{R}$ obtained using the OpenPose keypoints, and compared with the ones provided with the additional ground-truth data. The error is computed as the absolute value of the angle differences (in degrees), that is 
\[ 
(e_\alpha, e_\beta, e_\gamma) = ( \left |  \alpha_{gt} - \alpha_{op}  \right |, \left |  \beta_{gt} - \beta_{op}  \right |, \left |  \gamma_{gt} - \gamma_{op}  \right | ) 
\] 
and averaged across all the images. The average error obtained is ($e_\alpha$, $e_\beta$, $e_\gamma$) = (8.98 , 7.16 , 6.62 ). To the best of our knowledge, the only previous work reporting head pose estimation results in the thermal domain is the one of Yu \textit{et al.}~\cite{yu2010head}, that estimates the head pose with an error $< 10^{\circ}$ only in 17\% of the cases. Kopaczka \textit{et al.}~\cite{kopaczka2018combined} also developed a pose estimation module for thermal imagery, but evaluate their classifier on sequences in the visible spectrum. However, recent state-of-the-art algorithms that estimate the pose from single RGB images obtain a general error around 5\textdegree\ in challenging ``in the wild'' scenarios, \textit{e.g.}~\cite{yang2019fsa}. Overall our method is rather accurate, and can be at the very least used in practical scenarios to detect whether a subject is facing the camera or not.

\subsection{Warmest face point ambiguity}\label{subsec:warm-point-amb-exp}
Thermal screening standards suggest to perform the temperature measurement in correspondence of the eye canthus as it is the region that best approximates the internal body temperature, thus being the warmest face area. Hence, one could argue that detecting the face and then searching for the hottest point could suffice to recover their location. We instead empirically found that this strategy would lead to many wrong detections. Some qualitative examples of detection errors are shown in Fig.~\ref{fig:misdetection}. This strategy often fails because, first, the skin surface is sensible to environmental conditions and can significantly change its temperature in a short time frame; additionally, head pose changes can also alter the measurement~\cite{vardasca2017influence}, as well as wearing accessories. In Fig.~\ref{fig:misdetection}~(c) we also tried restricting the search space to the eyes region, using the manually annotated landmarks to ensure a correct localization of the eyes. Also in this case, such search can result in wrong detections, often when the subject is not correctly facing the camera. This result suggests that: (i) accurately localizing the eye canthus region is important to prevent detecting a wrong measurement point; (ii) it is likewise important to detect when the subject is not facing the camera. 

\begin{table}[!t]
\renewcommand{\arraystretch}{1.3}
\caption{Execution Time of each step of the algorithm (FPS / Sec)}
\label{tab:exec-time}
\centering
\begin{tabular}{|c|c|c|c|c|}
\hline
OpenPose & 3D Pose & Visiblity & Refinement & Tot\\
\hline
22 / 0.04 & 1K / 0.001 & 22 / 0.04 & 33 / 0.03 & 9 / 0.11\\
\hline
\end{tabular}
\end{table}

\subsection{Computational Time}\label{subsec:comput-time-exp}
Our approach is composed of three main steps, for which we report the execution time separately in Table~\ref{tab:exec-time}. The most computationally onerous steps are the OpenPose detection and the visibility mask computation, which both run at approximately 22 FPS (on a GTX 1080-Ti GPU, and Core i-7 CPU, respectively). The whole pipeline runs at approximately 9 FPS. Note that this is a prototype evaluation, and further optimizations can speed up the process. 

In practical screening scenarios, time constraints are strict, and computing a per-point visibility mask might not be strictly necessary. In such cases, one can either choose to estimate the face pose and advise when the subject is not frontal \textit{e.g.} fixing a threshold, or pre-compute an arbitrary number of masks corresponding to a discretized and fixed set of head rotations. This way the computational burden is noticeably reduced. 

\section{Limitations and Future Work}\label{sec:limitation}
The proposed approach can fairly accurately locate the eye canthus by refining the detections of an off-the-shelf algorithm. Still, there are some aspects that need further consideration. For example, when a subject wears eyeglasses, eyes are hidden by a black region; this happens because infrared sensors cannot measure the temperature on reflective surfaces. If the detector ( OpenPose in this case ) is robust enough to return an estimate of the eyes position, then our approach still will provide an approximate location of the canthus. In this scenario, the temperature measurement would be invalid (the temperature would be out of a reasonable range) and an alarm can be raised. However, the best solution would be that of detecting the presence of glasses a priori with a glasses detector, and we will address this issue in future developments. 

\begin{figure}[!t]
\centering
\includegraphics[width=0.99\linewidth]{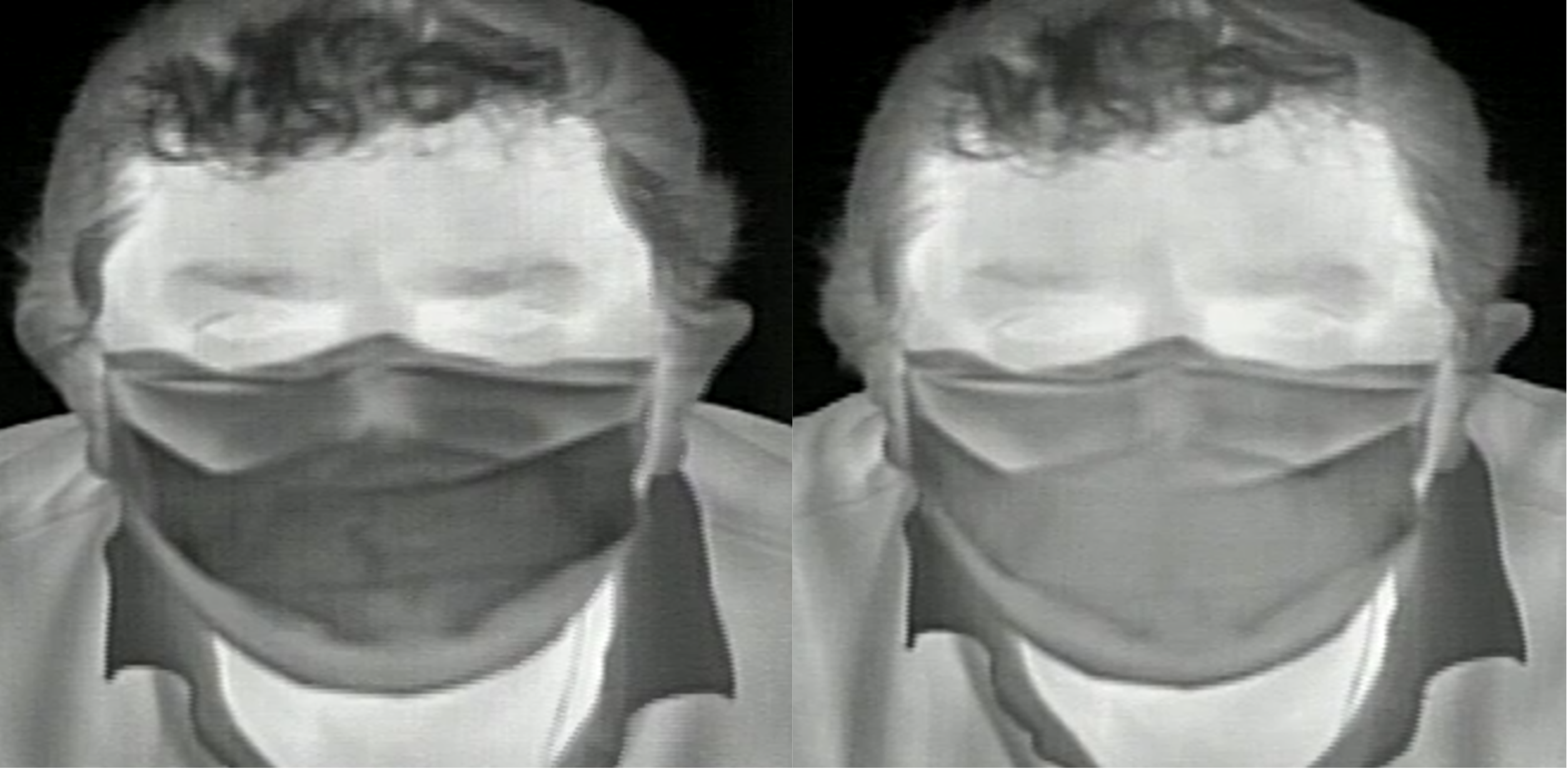}
\hspace{2cm}
Inhaling
\hspace{2.5cm}
Exhaling
\caption{Example of temperature dynamics when wearing a surgical mask.}
\label{fig:mask}
\end{figure}

Another factor that should not be ignored is the use of surgical masks. During a pandemic, wearing a mask may be a strict requirement in crowded or indoor places so this aspect should not be disregarded. In fact, we observed that wearing a surgical mask, differently from other accessories like hats or scarfs, can alter the temperature that is measured at the eye canthus (see Fig.~\ref{fig:mask}). This happens because cold/hot air flux generated when inhaling/exhaling or talking vents out directly around the eyes, and implies a strong temperature dynamics. This suggests collecting the measurement considering a short time frame, and that safety protocols should take into account this fact.  

Finally, we couldn't perform real temperature measurements as no dataset is publicly available that contain per-point temperature information. In the future, we aim at collecting a dataset including such information for carrying a deepened analysis and validation of our approach.

\section{Conclusion}\label{sec:conclusions}
In this paper, we have proposed an approach for detecting the eye canthus regions for human body temperature screening. Non-intrusive and non-contact thermography represents a viable and effective solution for mass screening of individuals for elevated body temperature, a common precursor of diseases. Our proposed approach works without employing images in the visible spectrum, thus preventing privacy related problems and allowing to use thermal-only cameras that have a reduced cost w.r.t.~bi-spectrum cameras. Our approach can accurately detect the eye canthi also in presence of facial expression and head rotations. In addition, it can reliably detect when the canthus regions are self-occluded and when the subject is not facing the camera, which can impair the temperature measurement. Furthermore, we provided additional annotations that are publicly released to an existing dataset, so as to ease future evaluations.

\section*{Acknowledgement}
This research was partially funded by Leonardo.






%
{\small
	\bibliographystyle{./IEEEtran}
	\bibliography{icprTherm}
}

\end{document}